\documentclass[3p,accept]{elsarticle}
\usepackage{times}
\usepackage{url}
\usepackage{lmodern}
\usepackage{comment}
\usepackage{listings}
\usepackage{tikz}
\usepackage{graphicx}
\usepackage{float}
\usepackage{hyperref}
\usepackage{multirow}
\usepackage{subfig}
\usepackage[normalem]{ulem}
\usepackage[most]{tcolorbox}
\tcbuselibrary{breakable}
\tcbset{
  promptbox/.style={
    colback=blue!5!white,
    colframe=blue!50!black,
    fonttitle=\bfseries,
    boxrule=0.5pt,
    arc=2pt,
    left=2mm, right=2mm, top=1mm, bottom=1mm,
    breakable
  }
}
\usepackage{xcolor}
\journal{Journal of \LaTeX\ Templates}

\begin{document}

\title{An integrated process for design and control of lunar robotics using AI and simulation}
\author[mymainaddress]{Daniel Lindmark}
\ead{daniel.lindmark@algoryx.com}
\author[mymainaddress]{Jonas Andersson}
\author[mymainaddress]{Kenneth Bodin}
\author[mymainaddress]{Tora Bodin}
\author[mymainaddress]{Hugo Börjesson}
\author[mymainaddress]{Fredrik Nordfeldth}
\author[mymainaddress,mysecondaryaddress]{Martin Servin}
\ead{martin.servin@umu.se}

\address[mymainaddress]{Algoryx Simulation AB, Ume\aa, Sweden}
\address[mysecondaryaddress]{Ume\aa\ University, Ume\aa, Sweden}

\begin{abstract}
We envision an integrated process for developing lunar construction equipment, where physical design and control are explored in parallel. In this paper, we describe a technical framework that supports this process. It relies on OpenPLX, a readable/writable declarative language that links CAD-models and autonomous systems to high-fidelity, real-time 3D simulations of contacting multibody dynamics, machine regolith interaction forces, and non-ideal sensors.

To demonstrate its capabilities, we present two case studies, including an autonomous lunar rover that combines a vision–language model for navigation with a reinforcement learning–based control policy for locomotion.

\end{abstract}

\maketitle

% Paper disposition                                                 pages
% Title, authors and abstract                                       0.25
% Introduction                                                      1 
% Related work                                                      0.5
% Background                                                        1.5 
%   - simulation of lunar construction robots
%   - motion control
%   - unified modeling with OpenPLX
% Integrated process                                                1.5 
%   - workflow and tools that enable an integrated approach 
% Results                                                           2
%   - scenario
%   - tests and output results
%   - observations from using the tools and workflow
% Conclusion, Acknowledgements                                      0.25
% References                                                        0.75
%                                                                   = 7.75

\section{Introduction}

Establishing a lunar base entails excavation and transport of regolith, on-site fabrication of construction elements, and their assembly into protective structures around surface modules deployed from Earth \cite{doi:10.1061/9780784484470.072}.
This requires mobile machinery capable of traversing the lunar surface and physically manipulating the environment. The design of such equipment and the development of its control systems each pose significant challenges. We advocate an integrated process where mechanical design and control algorithms are developed hand-in-hand, with the overall system architecture and performance as guiding principles. Our process and framework seamlessly connect CAD with high-fidelity, physics-based 3D simulation, creating a unified environment for designing, training, and validating autonomy, including perception, planning, and control.

It is widely recognised that achieving high-performance mobile robotic systems requires an integrated, modular design process~\cite{Allison2014}. Likewise, the importance of multidisciplinary modular design optimization has long been established in space engineering~\cite{sobieski1993multi}. However, in practice, development occurs mostly sequentially. First, the physical design space is explored, delimited by previous experience and proven principles, using simple control algorithms. The development of refined control algorithms and the selection and placement of sensors, usually occur when the physical design is fixed. This simplifies the steps but greatly reduces the solution space and generally leads to suboptimal results. An integrated process is challenging for both organisational and technical reasons. It requires sophisticated simulators that can accurately represent the dynamics of the full system while offering freedom to explore both the physical system and the control system design spaces. Exploring and searching for near-optimal solutions in this high-dimensional design space may be very computationally demanding. The integrated process has been given a formal mathematical description \cite{Allison2014,Herber2019}, and there are examples of successful use of algorithms such as nested co-design \cite{Herber2019} and Bayesian morphology-control co-optimization process \cite{Rosendo2017}.

Testing many different designs and control algorithms in a simulated environment is not only computationally demanding. It involves a large amount of tedious modeling of the many variations in design and scenarios. If done manually, this becomes a limiting factor. Therefore, automated support for this is starting to emerge. Recent work includes the generation of simulation scenes, in the form of URDF files, from images \cite{Chen2024urdformer} and simulatable models of quadrupedal robots from user text specifications and performance preferences \cite{Ringel2024}.

We present a technical framework that supports an integrated process to develop the design and control of lunar robots. At its core are the open-source physics modeling language OpenPLX and the AGX Dynamics physics engine. On top of this, we integrate several toolboxes which enable teams of specialists across machine design, robotics, control, and AI to work together to design, test, and validate new concepts in realistic lunar scenarios. 
The framework has been adopted in several studies and projects with the Swedish National Space Agency (SNSA), the European Space Agency (ESA), and the Japan Aerospace Exploration Agency (JAXA). The study \cite{linde2025simulation} was commissioned from Komatsu Ltd, which 
commissioned for the Space construction innovation project as
part of the Stardust Program in Japan.

As a demonstration of the framework we present two case studies, where the framework has been used to design and simulate an autonomous system on the Moon. The first is a novel scenario, which involves a six-wheeled rover navigating to targets specified in natural-language using a vision-language model (VLM), reinforcement learning, and low-level controllers. The second summarizes a previously published work \cite{linde2025simulation}, which highlights collaboration between an excavator and a dump truck. Together, these studies showcase the framework's versatility and potential for high-fidelity simulation for iterative development and validation of complex autonomous behaviours in realistic lunar environments.

%-------------------------------------------------------------
\section{Background}
This section introduces the underlying theory and methods on which our framework rests.

\subsection{Simulation of lunar construction robots}
For physics-based simulation of autonomously
controlled construction robots interacting with dynamic terrain, 
we rely on the physics engine AGX Dynamics \cite{AGX2025}.
AGX is based on the framework for contacting multibody dynamics introduced in
\cite{Lacoursiere2007}, extended to discrete element models (DEM) in \cite{Servin2014}, and to terrain deformation dynamics in \cite{servin2021multiscale}. Specifically, the simulation framework uses maximal coordinate representation
in terms of particles, rigid bodies, kinematic constraints, and complementarity conditions for joints, motors, and frictional contacts.
This has been used in previous work to develop machine learning models for autonomous
control of earthmoving equipment \cite{backman2021continuous,aoshima2023predictor} and 
active suspension system for rough terrain navigation \cite{wiberg2021control} with successful transfer of 
controllers trained in simulation to reality \cite{wiberg2024sim}.

The time-stepper and solver are optimized particularly for fixed timestep
realtime simulation of multibody systems with non-ideal constraints and non-smooth dynamics \cite{lacoursiere2011}. 
The modeling framework supports strong multiphysics coupling
that includes flexible bodies, hydraulics, hydrodynamics, drivelines, and more.
The novel hybrid direct-iterative solver supports fast and stable simulations at 
high accuracy for mathematically stiff and ill-conditioned systems \cite{lacoursiere2011},
such as robots and vehicles, 
and scalability to large-scale dynamic contact networks, 
such as for granular systems, at the price of lower accuracy \cite{Servin2014}.

For deformable terrain, AGX uses the multiscale model described in
\cite{servin2021multiscale}. It has been demonstrated to produce digging forces 
and soil displacements with an accuracy comparable to that of a resolved discrete 
element method (DEM), coupled with multibody dynamics, and with field tests involving 
full-scale construction equipment \cite{aoshima2024examining}.
A wide range of soil properties is supported. These are
parametrized mainly by their bulk mechanical properties,
including internal friction, cohesion, dilatancy, mass density, 
and compression index.

\subsubsection{Mobile robots and construction equipment}
The AGX Dynamics physics engine includes modules for high-level modeling of robots, vehicles, and construction equipment. This includes a drivetrain library with different motors, shaft, clutches, gears, and differentials, wheels and tracks modules with different levels of fidelity.
A body can be assigned the property of a  \emph{wheel} or a \emph{digging tool} that then interacts with the deformable terrain to accomplish bulldozing, excavation, or carrying soil. The reaction forces and soil displacements depend on the assigned soil properties. 

\subsubsection{Sensors}

Accurate physics simulation is necessary but not sufficient for transferring control systems from simulation to real-world deployment. Since real control systems rely on sensor feedback, the simulation environment must also represent non-ideal sensor behavior.

In this work, we employ AGX Dynamics, which provides simulation models for a range of sensors, including LiDAR, RGB-D cameras, inertial measurement units (IMUs), position encoders, and odometers. Predefined models of specific sensor types, such as Ouster LiDARs, reproduce essential characteristics including resolution and ray patterns.

All sensor models are configurable to incorporate disturbances and noise, allowing systematic investigation of robustness under imperfect sensing conditions. Such capabilities are critical for addressing the sim-to-real gap and for evaluating both learning-based and classical control methods in realistic scenarios.

\subsubsection{Lunar terrain}
Meteorite impacts have formed the lunar surface with steep, deep craters that the sun never reaches, crater rims along the poles that the sun almost always shines on and covered it all with a several-meter-thick layer of heterogeneous regolith. Traversing this regolith covered landscape is very challenging due to its physical properties. Properties which can also differ significantly based on location, depth and temperature. The AGX deformable terrain model has been parameterized to simulate lunar regolith, using parameters derived from existing literature \cite{slyuta2014physical}. This parameterized model has been applied in the project \cite{linde2025simulation}.

The Lunar Reconnaissance Orbiter has mapped the Moon’s surface since 2009, and there are several works using the data to create a detailed Lunar Digital Elevation Model (LDEM) of the surface, for example \cite{BARKER2021105119, BARKER2016346}. Within our framework, we will offer models of key lunar sites. While exact digital twins are impossible due to LDEM resolution (5m/pixel at best), we'll create realistic versions using algorithmic upscaling and domain randomization for both LDEM and regolith properties. Which can be used for sensitivity analysis of the autonomy stacks performance and for training robust AI-based controllers. 

\subsection{Motion Control}
Motion control enables robots and machines to move predictably and precisely to accomplish specific tasks. Optimal motion control is tightly coupled with the dynamical model of a robot. AGX Dynamics utilizes its existing knowledge of the dynamical model to offer automated support for motion control methods, e.g. forward/inverse kinematics, and computed force (inverse) dynamics. In computed force dynamics, AGX Dynamics uses the existing dynamical model of a robot and solves for the forces that are required to reach the next planned positions and velocities. Pairing a simpler feedback controller with computed force controllers results in a full-body controller that also can handle missing aspects of the model or unforeseen events. When concurrently designing the robot and the control system, the automatic coupling between the robot's model and its motion control is essential.

Additionally, AGX Dynamics have support for interfacing with third party robot motion control libraries and frameworks. Either directly through the API or using network communication. It natively supports ROS2 subscribers and publishers for standard message types, enabling use of the ROS2 ecosystem.

\subsection{Unified modeling with OpenPLX}
OpenPLX \cite{OpenPLX2025} is a - for humans and AI-agents - readable and writable declarative language for defining models, scenarios, and configurations using composable components. It builds on core bundles for math and physics, which are mapped once to a simulation runtime. Higher-level models, such as robotics or heavy machinery, can then be simulated without further mapping, enabling scalable and reusable simulations.  OpenPLX goes beyond competing modeling formats like USD and URDF by natively supporting arithmetic expressions, cross-file references, and external format import. Additionally, OpenPLX offers single inheritance and traits, allowing components to be efficiently specialized and composed. Machine parts are virtually assembled into templates of different machine categories , which can be extended or refined to produce testable variants. These features democratize the creation of configurable models for custom and domain-specific machines while preserving a single source of truth. With hierarchical and composable templates, OpenPLX enables rapid iteration and optimization in virtual environments. 

To ensure that an OpenPLX model accurately reflects its real-world counterpart, both physical behavior and control interfaces must be considered. OpenPLX components include built-in signal interfaces that can be directly mapped to real-world systems, enabling seamless integration and testing.

\subsubsection{OpenPLX model representation}
An OpenPLX \textit{bundle} is a collection of such models organized to represent domain-specific model categories with built-in physics descriptions. For example, vehicle systems or robotic machinery. These declarative models are interpreted by the OpenPLX parser, which validates the structure, reports errors in incomplete models, and generates a hierarchical model tree.
Rigid bodies and other physical components may have, but do not require, explicitly defined positions. Undefined transforms are computed from the OpenPLX mate definitions. The Physics3D bundle includes an algorithm for automatic assembly of articulated systems — including those with closed loops — provided that the assembly order is unambiguous. This enables features such as domain randomization of initial configurations and optimization of structural design, where parameters like link lengths or bearing placements can vary between iterations.
Once a machine concept is defined using OpenPLX, it can be reused and extended for iterative development toward an optimized solution. In addition, material properties defined in the model can be validated and calibrated using parameter estimation techniques based on real-world observations.

%-------------------------------------------------------------
\section{Integrated process for design and control}
%The framework is designed to support iterative development of optimized solutions across a wide range of applications, such as machine design and autonomy for mobile systems in the context of establishing a lunar base. The process must support iteration across disciplines and workflows covering a large solution space, i.e. a larger solution space than when work is done sequentially.

We envision an integrated process that addresses the simultaneous development of lunar robots' physical design and their autonomy solutions. The process consists of multiple workflows. 

%Effective collaboration across disciplines is essential to support collaborative design, analysis, testing, validation, AI-training, and related activities. The process consists of multiple workflows. 

%These workflows are not performed sequentially, in a integrated process they are rather iterated on continuously, until a optimal solution has been found. Iteration creates a larger solution space than if each workflow is done sequentially, e.g first the machine is designed with actuators and sensors placed, then the control system for that machine developed. 

\begin{itemize}
    \item Machine building: create and modify CAD models, assign physics attributes, instrument with sensors and actuators
    \item World building: modeling terrains, objects, and their physical properties.
    \item Scenario design: automation tasks of different difficulty and variability
    \item Autonomy design: developing control and decision-making systems
    \item Parameter exploration and validation: parameter identification and domain randomization for minimizing the sim-to-real gap.
    \item Batch simulation management: efficient parallel execution on available computing resources.
    \item Analysis and learning: data post-processing, learning, analytics, and visualization.
\end{itemize}

These workflows should not be performed sequentially. Instead workflow execution should be iterative, as users (human or agentic) continuously learn from simulation outcomes, other workflows, or emerging ideas. Effective collaboration across specialist teams (multi-agents) is essential while iterating through the workflows, to support collaborative design, analysis, testing, validation, and training of AI models.

The iterations result in frequent revisions in models and autonomy solutions. Updates to one component often necessitate corresponding changes in dependent components, which may otherwise be invalidated. Reliable versioning is therefore crusial throughout the process. Certain updates can be automated. For example, changing the dimension of links in a robot can be automatically propagated to the forward and inverse kinematics and dynamics. But the controllers might need re-tuning and re-training. Adding or removing sensors and actuators might call for revising the control strategy altogether. Keeping track of these dependencies are essential to make the corresponding changes in other workflows.

\subsection{A technical framework supporting the integrated process}
% We develop a technical framework, with software tools and interfaces, that address the various challenges for supporting the integrated process. The glue between the tools is OpenPLX and a versioned data store that ensures a single source of truth models that can reliably depend on each other. The framework is built using web and cloud technologies, which come with several benefits:

% \begin{itemize}
%     \item No installation required to get started in the browser
%     \item Hosting on-premise or in the cloud
%     \item Collaboration across teams
%     \item Enabling versioned data exchange API:s used by tools integrating with the framework.
% \end{itemize}
We are actively developing a technical framework, with software tools and interfaces, that addresses the challenges in the process outlined above. At the core of the framework are AGX Dynamics and OpenPLX together with a versioned data store that provides a single source of truth, ensuring that models can reliably depend on each other. On top of this, we are currently adding tools and GUI-elements, such as diagram editors and 3D-viewers, to support user workflows, effectively creating a one-to-one mapping between the human-readable OpenPLX models and the GUI-tools. In addition, we are working on AI agents that will be able to assist users in writing and modifying OpenPLX models, as well as working directly with the framework API to perform tasks. Making the AI-agent a collaborator across all workflows from the very beginning.

The framework is built on modern web and cloud technologies, which will bring several benefits, such as eliminating the need for installation by running directly in the browser, supporting both on-premise and cloud deployment, facilitating collaboration across teams, and enabling versioned data exchange APIs for tools that integrate with the framework.

%Additionally, a fully integrated AI-agent will assist users in writing and modifying OpenPLX models and utilizing the framework's API to perform tasks. This approach makes the agent a collaborator in the workflow from the very beginning. When completed, users will be able to provide complex compound instructions to define scenes, training scenarios, and quickly set up plots and analysis tools. This agent will also help to complete simple but tedious tasks, such as refactoring the model code, organizing diagram nodes, and setting up signal interfaces.

% Although GUI- and AI-based tools will enhance user workflows, it will remain possible to work directly with the OpenPLX language at every stage of the process. This ensures that machine models are consistently usable across all integrated frameworks and can also be modified through external LLMs.

\begin{figure}[htbp]
  \centering
  \includegraphics[width=1\linewidth]{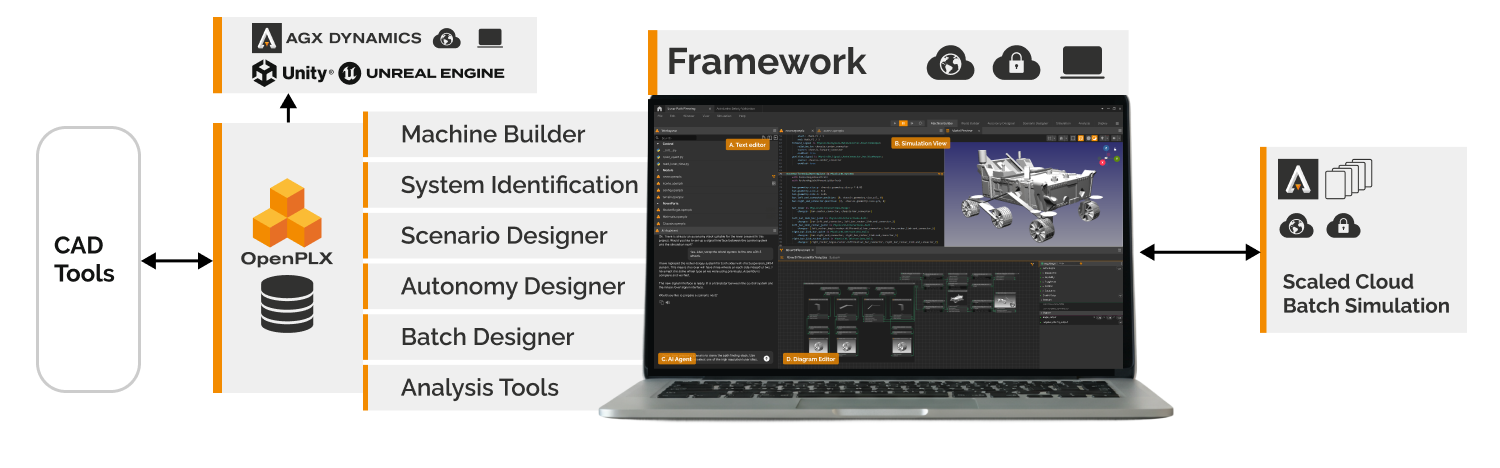}
  \caption{Illustration of the technical framework and connections to other tools.}
  \label{fig:platform-ecosystem}
\end{figure}
 
Figure \ref{fig:platform-ecosystem} illustrates the workflows within the framework and its connections to other tools. In what follows, we will elaborate on these workflows and how they interconnect. The design and tooling for some of the workflows are in an early mock-up phase, while others have prototype implementations.

\begin{figure*}[t]
  \centering
  \includegraphics[width=0.98\textwidth]{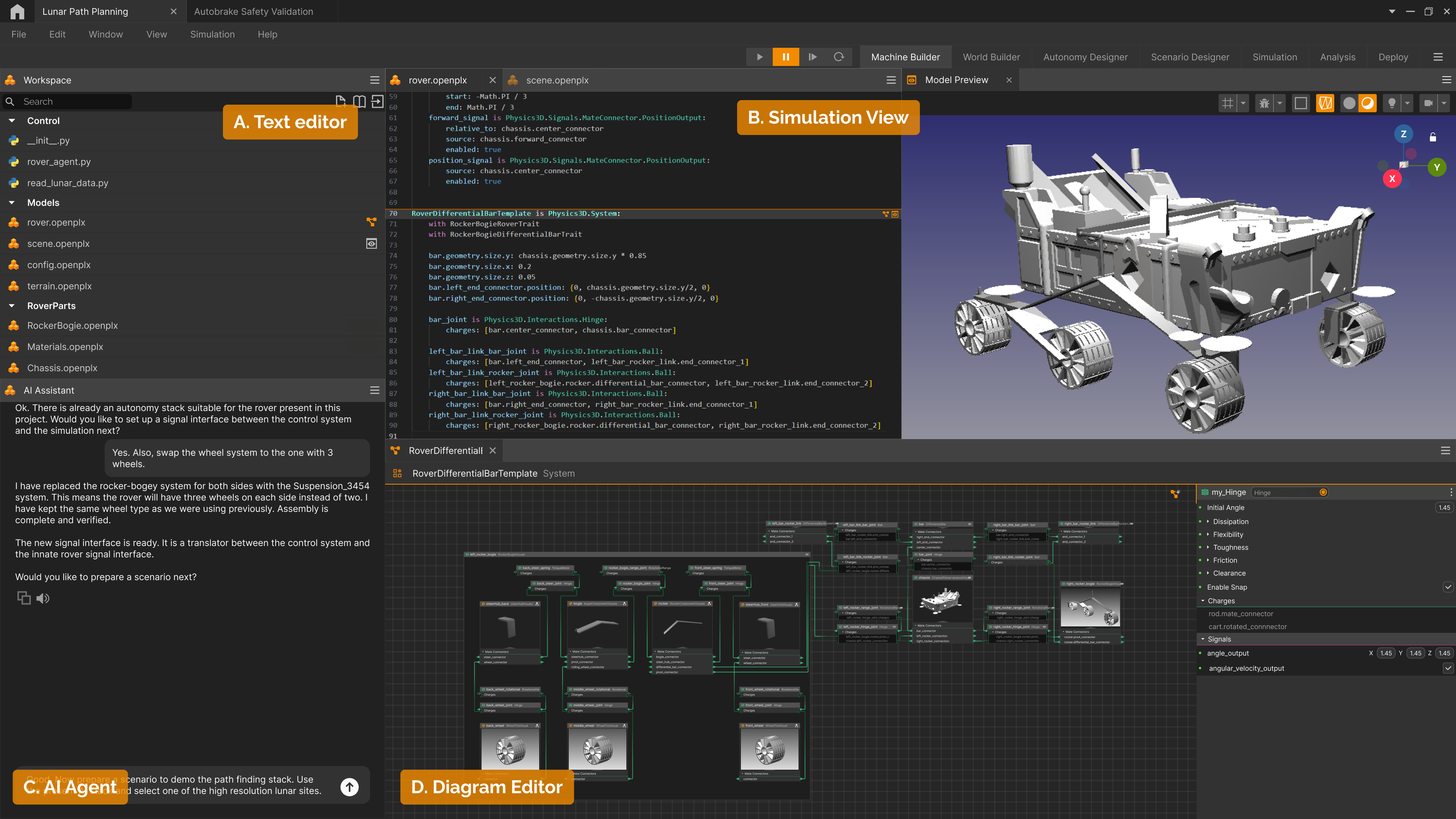}
  \caption{Example of how to interface with the framework with  (A) OpenPLX text editor, (B) Model 3D view, (C). integrated AI agent assistance and D. Diagram editor for no-code manipulation of OpenPLX models. }
  \label{fig:plx-tools}
\end{figure*}

\subsubsection{Machine Building Workflow}
The machine building workflow is focused on designing machines and preparing them for simulation. OpenPLX is a key factor of this workflow and serves as the bridge between external formats (.stl, urdf, obj, etc) and simulation by supporting import, modification, and debugging of pre-existing models. It will also contain tools for designing and comparing machine variants and for building new machines from scratch through different methods of AI-Assisted Design \citep{Elahi2023}. An OpenPLX bundle for high-level modeling of lunar robotics will enable users to model rovers, landers, robotic arms etc. using a higher level vocabulary. The bundle defines a library of reusable, high-level models and configurable components for assembly of machines and equipment, enabling rapid virtual prototyping of systems for lunar base construction and operations.

To support the machine building workflow, the framework includes an integrated OpenPLX text editor, a diagram editor and a 3D preview, as shown in Figure \ref{fig:plx-tools}. The text editor allows direct manipulation of OpenPLX models and the diagram editor functions as a low-code to no-code interface to primarily work with connecting and managing contact properties between machine- and scene components. This approach is intended to support users in quickly preparing modular systems and subsystems, as well as to lower the threshold of collaboration between experienced OpenPLX modellers and engineers working with other aspects of simulation.

\subsubsection{System identification}
The system identification workflow will enable selection and calibration of simulation models against empirical data. By importing sensor data from experiments and field operations, users can refine parameters such as stiffness, damping, friction etc. to ensure that simulated machines behave consistently with their physical counterparts.

\subsubsection{Scenario designer}
The Scenario Designer provides the ability to model lunar environments. Users will be able to import existing  LDEM terrain profiles and perform algorithmic upscaling and variations of these, or create a terrain profile suitable for their use-case. It will also be possible to change environmental conditions (e.g. angle of the sun), and create interaction events to create test cases tailored to specific operational contexts. Enabling repeatable evaluation of machine performance and autonomy solutions.

\subsubsection{Autonomy Design Workflow}
The design of an optimal autonomy solution, what parts it consists of and how they are connected, is both an application dependent problem but also an open research problem. Typically, there is a hierarchical structure to the control system. Where traditionally, a designed and task-specific state machine or behavior tree controls the sequence of executing different sub-modules, e.g. perception, sensor fusion, localization, mapping, prediction and  motion control. 

However, the introduction of foundation models is challenging how to design autonomous systems, how many levels are needed, and how much responsibility each level should carry, see for example \cite{kim2024openvlaopensourcevisionlanguageactionmodel, black2024pi0visionlanguageactionflowmodel, lbmtri2025}, which aims to replace almost the entire system, except the lower-level feedback controllers, with one single large AI-model. The robotic foundational models are typically transformer-based LLMs or VLM pre-trained on large scale internet data and trained to predict action sequences, thus called vision language action models (VLA). They are then fine-tuned to output robot actions, typically end-effector or joint, positions or velocities, based on natural language instructions and image feed. They have great potential for generalization across tasks and robots, but also introduces new safety challenges and requires a lot of computation close to the robot.

We support development and testing across the full spectrum of autonomy approaches. The OpenPLX signal interface provides seamless communication between autonomy components and a robot’s sensors and actuators, whether via API calls or standard communication protocols. The runtime environment allows Python scripting, enabling extensive customization and integration with both open-source and proprietary libraries, including pre-trained AI models. It also enables the collection of simulation data for fine-tuning or training AI models from scratch, and supports custom environments for reinforcement learning policy training.

%-------------------------------------------------------------
\section{Case Studies}
To demonstrate the applicability and versatility of the proposed simulation framework, we present two case studies. The first describes a novel scenario, while the second summarizes a previous published study \cite{linde2025simulation}. Together, these examples show the framework's breadth in modeling and simulating various lunar relevant machines and interactions, as well as in the possible ways of designing their autonomy systems.

\subsection{Case study 1 - rover navigation task}

\begin{figure}[htbp]
  \centering
  \includegraphics[width=\linewidth]{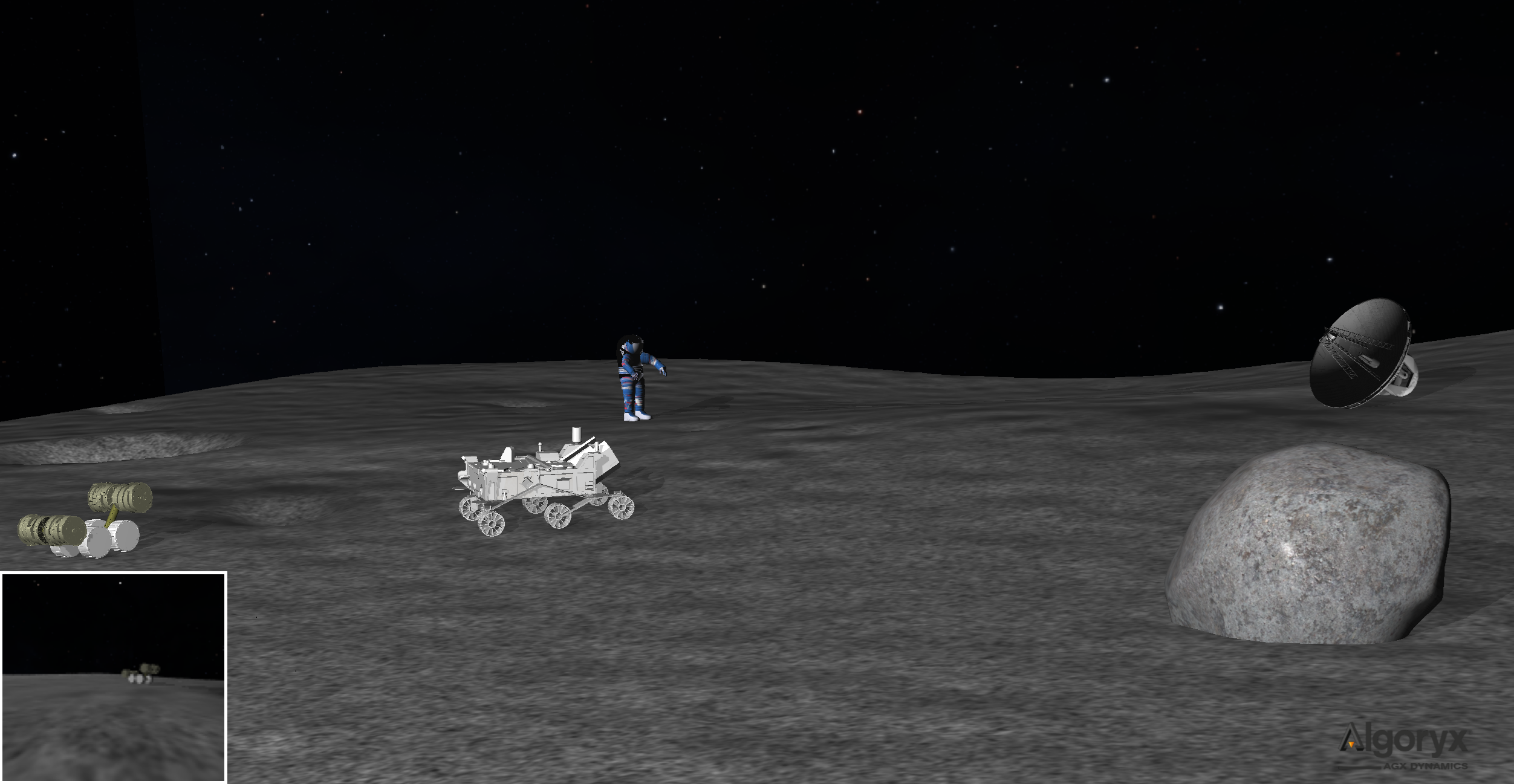}
  \caption{Simulation scene from Case study 1 with an autonomous lunar rover navigating with the aid of a VLM, the rover's camera view is visible in the bottom left corner.}
  \label{fig:demonstration-scenario}
\end{figure}

Using the presented framework, we have designed a rover and its autonomy system to solve a navigation task set on the Moon. Figure \ref{fig:demonstration-scenario} shows the running scenario. The ground is relatively flat, with the occasional small, shallow craters. Surrounding the rover there is a parabolic antenna, an astronaut, a larger rock, and NASAs Rassor excavator rover \cite{RASSOR2016}.

The rover's six wheels have individual speed control, and the four corner wheels are individually steered. The wheels are mounted on a differential bar rocker-bogie suspension system. The rocker-bogie suspension, together with the wheel joints, sums to 20 constraints, whereof 10 are actuated. 

\begin{figure}[htbp]
  \centering
  \includegraphics[width=0.5\linewidth]{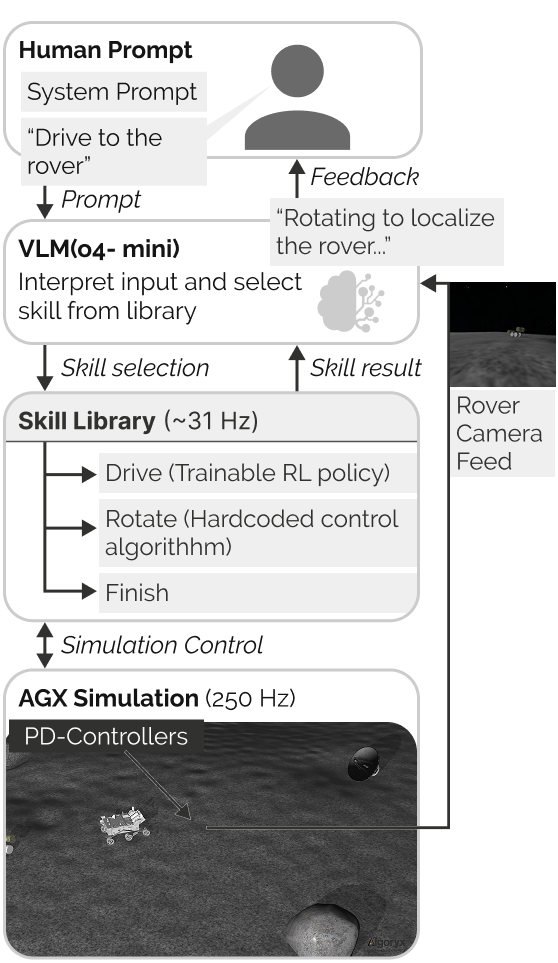}
  \caption{The lunar rover autonomy system. }
  \label{fig:autonomy-stack}
\end{figure}

We explore a hierarchical autonomy system that combines a VLM (OpenAI's o4-mini), a skill library (hardcoded and RL-trained skills), Ackermann steering, and PD-controllers (Figure \ref{fig:autonomy-stack}). The VLM's system prompt details targets, executable skills, and response format. A human operator provides high-level navigation targets in natural language, the VLM then interprets the task, selects skills based on the rover camera's ego view, determines task completion, and awaits new instructions.

The Drive RL-policy, trained with stable-baselines3's PPO, uses a 128x128 RGB image, wheel angles, forward velocity, and previous action (non-visual observations stacked twice) as observation. The action space consists of two values: Ackermann steering radius and target drive speed, which are converted to targets for the wheels individual steering angle and rotational speed. The reward function considers heading angle and target proximity with penalty for non-smooth actions and task duration. The rover is also supplied with an explicitly programmed skill of 60 degrees pivot turning.

% \[
% r = c_d(\Delta d) + c_a(1-40\Delta \theta^2) - r_t - c_s(a_{t-1} - a_{t})^2
% \]

Supplementary material can be found at \url{https://www.algoryx.com/papers/astra25-lunar-robotics}. It includes demonstration videos illustrating the system's behavior, together with an appendix that documents the system prompt, hyperparameters and experimental configurations employed during RL training.

\subsubsection{Results and observations from using tools and workflow}
The rover is able to solve short horizon tasks with a single target, i.e. \textit{"Drive to the large rock."} as well as longer horizon tasks such as, \textit{"Drive to the parabolic antenna, the rover, and finally to the astronaut"}, without intervention. It uses the \textbf{Rotate} skill until the target is in view and then activates the \textbf{Drive} skill. The system is able to recover from some types of errors. If the rover approaches the wrong target, the VLM notices this and tries to re-orient the rover towards the correct target again. If the instruction from the human is unambiguous, i.e. \textit{“Drive to a target”}, the VLM is able to ask for additional information to solve the task.

During the development of the controller, we frequently re-designed the rover and the environment, adjusting aspects like wheel geometry, rocker-bogie kinematics, sensor signals, and the terrain profile. OpenPLX simplified the implementations and version control of new designs. This made it easy to track what alterations were responsible for what improvements and failures. Rover re-designs typically necessitate re-training the RL control policy. Within the framework these dependencies are possible to follow, making it easy to move back and forth between designs for additional experiments and comparisons of behaviour or performance.

\subsection{Case study 2 - excavator and dumper collaboration task}

The framework was used in \cite{linde2025simulation} to design and evaluate an autonomy solution for an excavator and a dump truck collaborating on ground construction tasks on the Moon. We find it instructive to highlight the study here, as it shows different modeling, simulation and autonomy systems compared to the first case study presented.

\begin{figure}[htbp]
  \centering
  \includegraphics[width=0.7\linewidth]{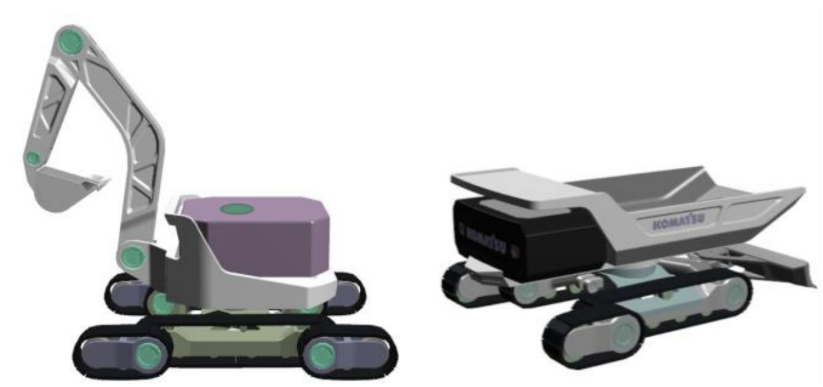}
  \caption{3D models of the excavator and the dump truck. Image taken from \cite{linde2025simulation}.}
  \label{fig:case2_excavator_and_dumptruck}
\end{figure}

The excavator and dump truck, shown in Figure \ref{fig:case2_excavator_and_dumptruck}, are each equipped with actively controlled articulated crawlers to enhance mobility on soft lunar regolith. Their crawler subsystems allow smooth traversal by lowering the sub-crawlers for reduced sinkage during excavation and raising them during turning manoeuvres to minimize resistance. The excavator is equipped with an articulated and actuated digging arm, while the dump truck has a tiltable truck bed for material unloading and an actively controlled blade for final grading. 

Behaviour Trees (BT) are used for high-level planning and for coordinating the machine collaboration. A BT organizes nodes in a tree structure, where the leaf nodes are actions that are performed and the parent nodes decides the traversal of the tree nodes depending on the return values of the leaf nodes, see \cite{IOVINO2022104096} for an in-depth description. 

\begin{figure}[H]
    \centering
    \includegraphics[width=0.75\textwidth,trim={140mm 75mm 140mm 80mm},clip]{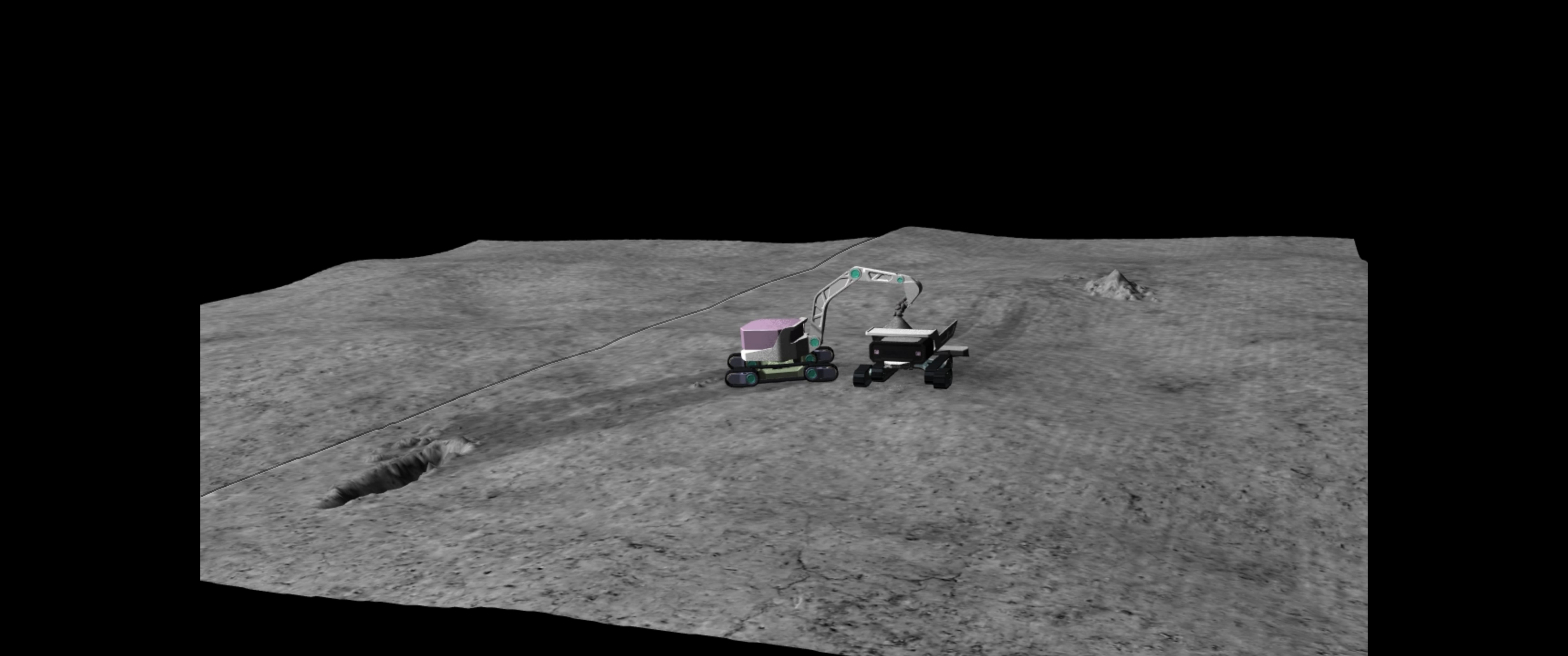}\\
    \vspace{1mm}
    \includegraphics[width=0.75\textwidth,trim={0mm 75mm 0mm 0mm},clip]{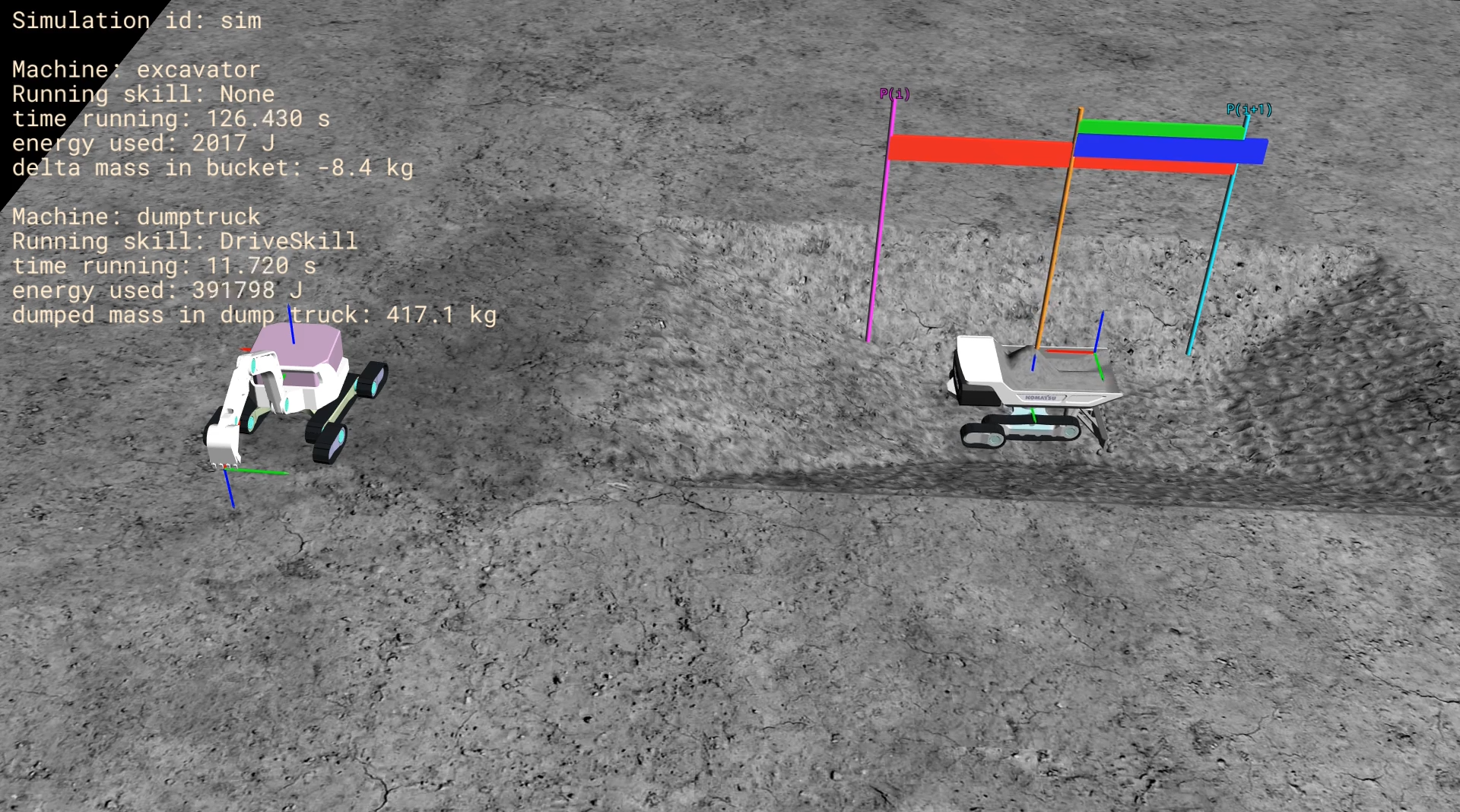}
    \caption{The top image shows the excavator loading regolith into the dump truck in the \textit{extracting lunar regolith} scenario. The bottom image shows the dump truck performing final grading in the \textit{excavating a predefined ground structure}. Both images are taken from \cite{linde2025simulation}.}
    \label{fig:case2_both_scnarios}
\end{figure}

\subsubsection{Results and observations from using tools and workflow}
Two long-running test scenarios are analysed, namely \textit{extracting lunar regolith} and \textit{excavating a predefined ground structure}. The top image in figure \ref{fig:case2_both_scnarios} shows the former scenario and the bottom the latter. For the \textit{extracting lunar regolith} scenario, they make a construction performance comparison, over 30 digging cycles, between operating on flat terrain versus a sloped terrain. They track the evolution of excavated and dumped mass, cycle time and work done per cycle. And can draw conclusions that there is room for improving the planning in collaboration between the two machines, and that aside from the crawlers the excavator arm actuator dominates the power consumption, among other things.

In \textit{excavating a predefined ground structure}, the excavator successfully excavated the desired terrain shape despite soil frequently avalanching into it, and the dump truck completed the final grading with a levelled surface that closely matched the 3D specification.

%-------------------------------------------------------------
\section{Conclusion}
Development of design and autonomy solutions for lunar robotics require an integrated process with full-system simulation of machines in dynamic environments, with many types of sensors, actuators, and controllers. Extended to the lunar domain, OpenPLX hierarchical and compositional structure enables management and coordination of the complex simulation infrastructure and processes involved in developing well-designed and autonomous systems. 
The case studies highlight the framework's versatility in modeling and simulation, testing variations in the machines physical design and autonomous control. 

\section*{Acknowledgments}
This research was supported in part by the Swedish National Space Agency through Rymdtillämpningsprogrammet (dnr 2024-00310, AILUR), by the European Space Agency (ESA) under the Discovery Programme through activity no. 4000146486 “Lunar Robotic Construction with Raw Regolith,” and by the European Commission through the XSCAVE project (HORIZON-CL4-2024-DIGITAL-EMERGING-01-03, RIA, Grant Agreement No. 101189836).

% \section*{References}
\bibliography{references}

% FROM THE TEMPLATE:
% The citations are either parenthetical \citep{smith96} or in-text
% as shown by \citet{allen73} and elaborated on by \citet{nobody97}.
%
% With \texttt{natbib}, in-text citations are generated with
% \verb!\citet{allen73}! to yield ``\citet{allen73}'' while parenthetical
% ones are made with \verb!\citep{smith96}! for \citep{smith96}. There are
% many other possibilities, such as \verb!\citeauthor! for the authors
% without year. See the \texttt{natbib} documentation.

% With \texttt{natbib} the bibliography must be entered differently, at
% least the \verb!\bibitem! entries.
% \begin{small}
% \begin{verbatim}
% \bibitem[Allen(1973)]{allen73}
% . . .
% \bibitem[Nobody et~al.(1997)]{nobody97}
% . . .
% \bibitem[Smith \& Jones(1996)]{smith96}
% . . .
% \end{verbatim}
% \end{small}
% The text in square brackets contains the author and year information,
% with the year part in parentheses, no space before, which is used by the
% \verb!\citet! and \verb!\citep! commands.

\appendix
\onecolumn

\lstset{
  basicstyle=\ttfamily\small,
  breaklines=true,
  breakatwhitespace=false,
  columns=fullflexible,      % preserves spacing/indentation
  breakautoindent=true,      % align wrap with the original line’s indent
  breakindent=0pt,           % no extra indent beyond the line’s own indent
  % Optional but nice: show a wrap arrow and ensure margins line up
  % postbreak=\mbox{\textcolor{gray}{$\hookrightarrow$}\space},
  % numbers=left,              % if you use line numbers...
  % numbersep=6pt,
  xleftmargin=2em,           % content left margin
  frame=single,
  framexleftmargin=2em       % match frame margin to content margin
}

\section{System prompt}
The system prompt used with the OpenAI \texttt{o4-mini-2025-04-16} model, with reasoning effort set to low, is reproduced below. It specifies the scene, defines the available skills, and prescribes the expected response format of the VLM.

\begin{lstlisting}[basicstyle=\ttfamily\small, breaklines=true]
You are a high-level lunar rover controller. You will get the current image view from the rovers perspective and the target described in text. The rover is placed on the lunar surface in the middle of a lunar base. Surrounding you there is a number of targets, an astronaut in a blue spacesuit, a round grey parabolic antenna reflector dish, another four wheeled rover and a larger rock. All of the targets are spread out meaning that two targets cannot be in view at the same time if they are not far away. Given the rovers camera view and the user given navigation task you will choose a skill from the skill library and give the necessary arguments for that skill.

The targets that the user can choose from are.
 - Astronaut
 - Antenna
 - Rover
 - Rock

The skill library:
- Drive(target) # Choose this skill when the target is visible in the image. Where the target argument is one of the possible targets listed above, without the quotation marks. The rover will then try to drive towards it for a set amount of time. Returns "Success" if it has reached the target and "Fail" if it did not. If the target is still in view and closer than before it failed due to time issues and you should choose the drive skill again.
- Rotate(target) # Choose this skill when the target is not visible, or if the view is dominated by the wrong target, or if the view is dominated by a large shadowed region. The rover will then rotate in place to try to locate the target. Where the target argument is one of the possible targets listed. Returns "Success" the if it managed to rotate. Returns "Fail" if it got stuck.
- Finish() # Choose this when the entire task provided by the user has been executed.
- Shutdown() # Choose this when user says it is done.
- MoreInformation() # Choose this when you need more information about the target

Your answer by first shortly describing the scene, what you intend to do and why. And then finish by choosing one skill enclosed by the skill tags <skill></skill>. If the user has not given you enough information to choose a target, you simply answer for more information about the target and choose the MoreInformation skill using <skill>MoreInformation()</skill>, without giving a specific target.

When the rover has executed a skill you will get a return value from that skill that is either, "Success" or "Fail", depending if the skill failed or succeded. You will also get the update camera view from the rover.


If you think that you are close enough to the target given by the user choose the <skill>Finish()</skill> skill. And ask for new instructions.

At "Startup!" you should only present yourself and say that you are awaiting further instructions. You should not pick a skill to do.

At "Continuation!" your history from the previous task is cleared. Just as at "Startup!" you should ask for new instructions without picking a skill.

\end{lstlisting}

\newpage

\section{RL training configuration}
The following hyperparameters were used for training the drive skill policy PPO. 
The learning rate was scheduled to decay linearly from the maximum to zero.
Unless otherwise noted, all remaining hyperparameters correspond to the default settings of PPO in \texttt{stable-baselines3} (v2.6.0).
\lstset{
  basicstyle=\ttfamily\small,
  breaklines=true,
  columns=fullflexible,
  frame=single
}

\begin{lstlisting}
{
  "algorithm": "PPO",
  "policy": "MultiInputActorCriticPolicy",
  "learning_rate": 1.4e-04,
  "n_steps": 2048,
  "n_envs": 8,
  "batch_size": 4096,
  "ent_coef": 1.68e-06,
  "n_epochs": 10,
  "total_timesteps": 1.5e7
  }
}
\end{lstlisting}

The reward function for the driving skill is designed to encourage progress toward the target while promoting stable and purposeful motion. It is defined as

\begin{equation}
r_t =
\begin{cases}
c_d \Delta d + c_a \bigl(1 - 40\theta^2\bigr) - c_t - c_s (a_t - a_{t-1})^2, & \text{step reward (progress, alignment, smoothness)} \\[6pt]
1, & \text{if the rover stops within $4\,\text{m}$ of the target (success)} \\[6pt]
-0.5, & \text{if $\theta > 30^\circ$ (failure, episode terminated)} 
\end{cases}
\end{equation}

where $\Delta d$ denotes the change in distance to the target, $\theta$ is the angle between the rover’s heading and the direction to the target, and $a_t$ is the normalized action at time $t$. The first term rewards forward progress, the second favors alignment with the target direction, the third penalizes idleness, and the last discourages large changes between successive actions. The constants are set to $c_d = 1.0$, $c_a = 0.02$, $c_t = 0.01$, and $c_s = 0.01$.

In addition, the rover receives a large terminal reward of $r_t = 1$ when it comes to a stop within $4 \,\text{m}$ of the target, signaling successful completion of the task. Conversely, if the heading angle exceeds $30^\circ$, the episode is terminated early, and a reward of $r_t = -0.5$ is given.
\end{document}